\pdfoutput=1

\documentclass[10pt, a4paper]{article}
\usepackage{lrec2014}
\usepackage{graphicx}
\usepackage{url}
\usepackage{mathtools}
\usepackage{multirow}
\usepackage[utf8]{inputenc}
\usepackage{textcomp}
\usepackage{enumitem}

\renewcommand\thesection{\arabic{section}}
\renewcommand\thesubsection{\thesection.\arabic{subsection}}

\title{Extracting a bilingual semantic grammar from FrameNet-annotated corpora}

\name{Dana Dann\'{e}lls, Normunds Gruzitis}

\address{
Spr\r{a}kbanken, University of Gothenburg\\
Department of Computer Science and Engineering, University of Gothenburg\\
Institute of Mathematics and Computer Science, University of Latvia\\
dana.dannells@svenska.gu.se, normunds.gruzitis@\{cse.gu.se, lumii.lv\}
}

\abstract{
We present the creation of an English-Swedish FrameNet-based grammar in Grammatical Framework. The aim of this research is to make existing framenets computationally accessible for multilingual natural language applications via a common semantic grammar API, and to facilitate the porting of such grammar to other languages. In this paper, we describe the abstract syntax of the semantic grammar while focusing on its automatic extraction possibilities. We have extracted a shared abstract syntax from {\texttildelow}58,500 annotated sentences in Berkeley FrameNet (BFN) and {\texttildelow}3,500 annotated sentences in Swedish FrameNet (SweFN). The abstract syntax defines 769 frame-specific valence patterns that cover 77,8\% examples in BFN and 74,9\% in SweFN belonging to the shared set of 471 frames. As a side result, we provide a unified method for comparing semantic and syntactic valence patterns across framenets.
\newline
}

\begin{document}

\maketitleabstract

\section{Introduction}

Over the last decade, the exploitation of lexical-semantic resources like FrameNet \cite{RuppenhoferEtAl2010} has been the focus of attention for a range of NLP applications such as semantic parsing \cite{DasEtAl2013}, information extraction \cite{MoschittiEtAl2003} and natural language generation \cite{RothAndFrank2009}. FrameNet which is organised according to the principles of frame semantics \cite{Fillmore1985} is an attractive candidate for solving advanced NLP tasks because it provides a benchmark for representing a large amount of word senses and word usage patterns through the linguistic annotation of corpus examples.

As a result, the work on building FrameNet-like resources for languages other than English has emerged. Today there are available computationally oriented framenets for German, Japanese, Spanish \cite{Boas2009} and Swedish \cite{BorinEtAl2010}. Other initiatives exist for Chinese, Hebrew, Hindi, Italian, Latvian, Polish, Russian, as well as for other languages.

The development of the wide coverage Berkeley FrameNet (BFN) has been a labour intensive task \cite{FillmoreEtAl2003}. The effort of conducting a framenet for a new language is somewhat easier because the conceptual backbone of BFN can be shared among languages. However, although new framenets often leverage from BFN, they are mostly not aligned and unified at the lexical and grammatical level. Each framenet uses its own annotation format, grammatical types and functions, and word sense inventories. By integrating framenets not only at the lexical level but also at the grammatical level, one would gain access to a powerful computational multilingual NLP resource.

In this paper, we address the integration at the grammatical level, describing automatic extraction of the abstract syntax of a currently bilingual but potentially multilingual FrameNet-based grammar. To define the grammar, we use Grammatical Framework (GF), a formalism and a resource grammar library for implementing multilingual application grammars. While we focus on English and Swedish, the same approach can be applied for extracting and comparing frame semantic grammars for other languages.

This paper is structured as follows. In Section~\ref{sec:background}, we introduce BFN and Swedish FrameNet, and explain the role of GF. In Section~\ref{sec:methodology}, we describe and discuss our design of the abstract syntax of a FrameNet-based grammar. In Section~\ref{sec:experiments}, we present experiment series we carried out to compare the effects of various choices made in the grammar extraction, and evaluate the final results. We conclude the paper in Section~\ref{sec:conclusions}.

\section{Background}
\label{sec:background}

\subsection{Berkeley FrameNet (BFN)}
\label{ssec:bfn}

FrameNet \cite{FillmoreEtAl2003} is a unique lexical knowledge base that is based on the theory of frame semantics \cite{Fillmore1985}.\footnote{\url{https://framenet.icsi.berkeley.edu/}} According to this theory, a semantic frame which represents different cognitive scenarios consists of various semantic slots (roles) called \emph{frame elements}~(FE), and is evoked by a target word called \emph{lexical unit}~(LU). A semantic frame carries valence information about different syntactic realizations of FEs and about their semantic characteristics. Syntactic and semantic valence patterns are derived from FrameNet-annotated corpora.

Frames are described by two types of FEs: \emph{core elements} and \emph{non-core elements}. Core FEs uniquely characterize the frame and syntactically correspond to verb arguments, in contrast to non-core FEs (adjuncts) which can be instantiated in many other frames. As an example, consider the semantic frame $\mathrm{Desiring}$ given in Table~\ref{tab:FrameEng}.

\begin{table}
\tabcolsep 2pt
\begin{center}
\begin{small}
\begin{tabular}{ll}
\hline
\multicolumn{2}{c}{$\mathrm{Desiring}$}\\
\hline
Definition:
& An $\mathrm{Experiencer}$ desires that an $\mathrm{Event}$ occur. In\\
& some cases, the $\mathrm{Experiencer}$ is an active partici-\\
& pant in the $\mathrm{Event}$, and in such cases the $\mathrm{Event}$ it-\\
& self is often not mentioned, but rather some $\mathrm{Focal\_}$\\
& $\mathrm{participant}$ which is subordinately involved.\\
\hline
Core:
& $\mathrm{Event}$, $\mathrm{Experiencer}$, $\mathrm{Focal\_participant}$,\\
& $\mathrm{Location\_of\_Event}$\\
Non-core:
& $\mathrm{Cause}$, $\mathrm{Degree}$, $\mathrm{Duration}$, $\mathrm{Manner}$, $\mathrm{Place}$,\\
& $\mathrm{Purpose\_of\_Event}$, $\mathrm{Reason}$, $\mathrm{Role\_of\_focal\_}$\\
& $\mathrm{participant}$, $\mathrm{Time}$, $\mathrm{Time\_of\_Event}$\\
\hline
\end{tabular}
\end{small}
\caption{The definition and FE sets of the frame $\mathrm{Desiring}$.}
\label{tab:FrameEng}
\end{center}
\end{table}

The BFN version 1.5 defines {\textgreater}1,000 frames. Each frame is evoked by specific LUs. For example, some of the LUs evoking the frame $\mathrm{Desiring}$ are: \emph{ambition.n.6702, crave.v.6596, craving.n.6597, desire.v.6413, desire.n.6414, eager.a.6546, feel\_like.v.7430, want.v.6412, yearn.v.6599}, where the number is the sense identifier.

BFN provides information about the semantic and syntactic valence of LUs. For example, consider valence patterns for the verb \emph{to want} given in Table~\ref{tab:feBFN}, where the syntactic annotations include phrase types, e.g. noun phrase ($\mathrm{NP}$), prepositional phrase ($\mathrm{PP}$), verb phrase ($\mathrm{VP}$), and shallow grammatical functions: external argument ($\mathrm{Ext}$), first object ($\mathrm{Obj}$), general dependent ($\mathrm{Dep}$).

\begin{table}
\tabcolsep 4pt
\begin{center}
\begin{tabular}{cll}
\hline
Examples & \multicolumn{2}{l}{Valence patterns}\\
\hline
40 & $\mathrm{Event}$ & $\mathrm{Experiencer}$\\
(22) & {\tt VPto.Dep} & {\tt NP.Ext}\\
\hline
14 & $\mathrm{Experiencer}$ & $\mathrm{Focal\_participant}$ \\
(10) & {\tt NP.Ext} & {\tt NP.Obj} \\
(1) & {\tt PP[by].Dep} & {\tt NP.Ext} \\
\hline
\end{tabular}
\caption{Some semantic patterns and some of their syntactic realizations found in BFN for \emph{want.v} evoking $\mathrm{Desiring}$.}
\label{tab:feBFN}
\end{center}
\end{table}

In this paper, we consider only those frames for which there is at least one corpus example where the frame is evoked by a verb. BFN comprises 556 such verb frames that are evoked by {\texttildelow}3,200 LUs in {\textgreater}68,500 annotated sentences.

\subsection{Swedish FrameNet (SweFN)}
\label{ssec:Swefn}

Swedish FrameNet is being developed as a part of a large project at Språkbanken \cite{BorinEtAl2010}.\footnote{\url{http://spraakbanken.gu.se/swefn/}}
The resource has been expanded from BFN, and thus mostly uses the same inventory of frames and FEs. For example, the description of the frame $\mathrm{Desiring}$ shown in Table~\ref{tab:FrameEng} is the same also in SweFN. Note, however, that {\texttildelow}50 additional frames that are not used in BFN have been introduced in SweFN, and {\texttildelow}15 BFN frames have been modified.

LUs in SweFN are linked to SALDO, the Swedish association lexicon \cite{BorinEtAl2013}. Some LUs that evoke the frame $\mathrm{Desiring}$ are: \emph{känna\_för.vb.1} `to feel like', \emph{längta.vb.1} `to yearn', \emph{vilja.vb.1} `to want', \emph{åtrå.vb.1} `to desire', \emph{begärelse.nn.1} `wish', \emph{åtrå.nn.2} `desire', where the number is the sense identifier.

All sentences in SweFN are syntactically annotated with MaltParser \cite{NivreEtAl2007}. Table~\ref{tab:feSweFN} shows some semantic valence patterns and their syntactic realizations for the verb \emph{känna för}. The syntactic patterns include morpho-syntactic tags, e.g. noun ($\mathrm{NN}$), verb ($\mathrm{VB}$), infinite form ($\mathrm{INF}$), and dependency labels, e.g. verb group ($\mathrm{VG}$), subject ($\mathrm{SS}$), object ($\mathrm{OO}$).

\begin{table}
\tabcolsep 4pt
\begin{center}
\begin{tabular}{cll}
\hline
Examples & \multicolumn{2}{l}{Valence patterns}\\
\hline
1 & $\mathrm{Event}$ & $\mathrm{Experiencer}$\\
(1) & {\tt VB.INF.VG} & {\tt NN.SS}\\
\hline
2 & $\mathrm{Experiencer}$ & $\mathrm{Focal\_participant}$ \\
(2) & {\tt NN.SS} & {\tt NN.OO}\\
\hline
\end{tabular}
\caption{Semantic patterns and their syntactic realizations found in SweFN for \emph{känna\_för.vb} evoking $\mathrm{Desiring}$.}
\label{tab:feSweFN}
\end{center}
\end{table}

As of February 2014, SweFN contains {\textgreater}900 frames of which 638 are evoked by {\texttildelow}2,300 verb LUs in {\textgreater}3,700 annotated sentences.

\subsection{Grammatical Framework (GF)}
\label{ssec:gf}

The presented FrameNet-based grammar is being developed in GF, a categorial grammar formalism specialized for multilingual (parallel) grammars \cite{Ranta2004}. One of the key features of GF grammars is the division between an abstract syntax and concrete syntaxes. The abstract syntax defines the language-independent structure (the semantics) of an application grammar or a resource grammar library, while the concrete syntaxes define the syntactic and lexical realization of the abstract syntax for particular languages.

Remarkably, GF is not only a grammar formalism or a specialized functional programming language. It also provides a general-purpose resource grammar library (RGL) for nearly 30 languages that implement the same abstract syntax, a shared syntactic API \cite{Ranta2009}.\footnote{\url{http://www.grammaticalframework.org/}}
The use of the shared types and functions allows for rapid and rather flexible development of multilingual application grammars without the need of specifying low-level details like inflectional paradigms, agreement and word order.

In GF, features and constituents of phrases are stored in objects of record types, and functions are applied to such objects to construct phrase trees. In the abstract syntax, both argument types and the value type of a function are separated by right associative arrows, i.e. all functions are curried. For example, in the application grammar for an on-line store, a function $\mathrm{Wish}$ might be declared:

\begin{enumerate}
\item[] $\mathit{fun}$ $\mathrm{Wish : Person \to Product \to Status}$
\end{enumerate}

The above function returns a phrase of type $\mathrm{Status}$ that is computed from two arguments of types $\mathrm{Person}$ and $\mathrm{Product}$ respectively. The exact behaviour of the function is defined in concrete syntaxes. For example, in the English grammar (and similarly in Swedish), one could, first, specify that the types $\mathrm{Person}$ and $\mathrm{Product}$ map to the RGL type $\mathrm{NP}$ (noun phrase) and that $\mathrm{Status}$ maps to $\mathrm{Cl}$ (clause). Second, the following RGL constructors could be applied to define the linearization of the function $\mathrm{Wish}$:

\begin{enumerate}[noitemsep]
\item[] $\mathit{lincat}$ $\mathrm{Status = Cl}$
\item[] $\mathit{lincat}$ $\mathrm{Person, Product = NP}$
\end{enumerate}
\begin{enumerate}[noitemsep]
\item[] $\mathit{lin}$ $\mathrm{Wish}$ pers prod $\mathrm{=}$
\begin{enumerate}[itemsep=0pt,parsep=0pt,topsep=0pt,partopsep=0pt]
\item[] $\mathrm{mkCl}$ pers $\mathrm{(mkV2}$ $\mathrm{(mkV}$ ``want"$\mathrm{))}$ prod
\end{enumerate}
\end{enumerate}

Regarding FrameNet, its conceptual, language-independent layer (frames and frame elements) can be defined in an abstract syntax, while the language-specific lexical layer can be defined in concrete syntaxes. The current syntactic API of GF can be used for generalizing and unifying the grammatical types and constructions used in different framenets. The resulting FrameNet-based grammar, in turn, provides a frame semantic abstraction layer -- a semantic API -- to the syntactic RGL. Such approach has been proposed before \cite{GruzitisEtAl2012}, but this is the first attempt to implement it on a wide coverage.

\section{A FrameNet-based grammar in GF}
\label{sec:methodology}

\subsection{The abstract syntax design}
\label{ssec:gf-abstract}

To keep the representation simple, the current design of the abstract syntax is focused on natural language generation. The resulting grammar would not be appropriate e.g. for a semantic role labelling task, i.e., for parsing directly with the semantic grammar because at this stage we do not limit the set of verbs that can evoke a particular frame, and we do not limit the set of prepositions that can be used for a particular FE if it is realized as an adverbial modifier. However, the current design allows for using the resulting grammar for parsing indirectly via an application grammar that uses it as an API, specifying appropriate verbs and appropriately formed modifiers.

The grammar consists of three main modules: an FE module, a frame module and an LU module.

{\bf The FE module} lists all core and potentially non-core FEs that belong to the verb frames declaring them as semantic categories (types) in the grammar. Although the conception of FrameNet is that core FEs are unique to the frame, even though their names are not unique across frames, we do not make any semantic difference in this module -- FEs are implicitly made frame-specific in the frame module. We subcategorize FEs by different syntactic RGL types that can be used to realize them either in the same frame or across frames. The mapping to RGL types is specified in a shared concrete syntax (in a similar way that is illustrated in Section~\ref{ssec:gf}), but this information is also encoded by a suffix in each FE name to keep the names unique.

The distinction between core and non-core FEs is made by adding the prefix $\mathrm{Opt}$ to each non-core FE. The same FE name can appear with and without this prefix, if across different frames it is declared as both core and non-core.

The following example shows some FE declarations that are relevant to the frame $\mathrm{Desiring}$:

\begin{enumerate}[noitemsep]
\item[] $\mathit{cat}$ $\mathrm{Event\_VP}$
\item[] $\mathit{cat}$ $\mathrm{Experiencer\_NP}$
\item[] $\mathit{cat}$ $\mathrm{Focal\_participant\_NP}$
\item[] $\mathit{cat}$ $\mathrm{Focal\_participant\_Adv}$
\item[] $\mathit{cat}$ $\mathrm{Opt\_Degree\_Adv}$
\end{enumerate}

Note that $\mathrm{Experiencer\_NP}$ is used (in a different meaning), for instance, in frames $\mathrm{Emotion\_heat}$ and $\mathrm{Experiencer\_focus}$, while $\mathrm{Event}$ is used, for instance, as $\mathrm{Event\_Adv}$ in $\mathrm{Participation}$ and as $\mathrm{Event\_NP}$ in $\mathrm{Preventing}$. $\mathrm{Focal\_participant}$ is typically realized as an NP in $\mathrm{Desiring}$, but some verbs, e.g. \emph{to yearn}, require it as a prepositional phrase, hence this FE is subcategorized using the syntactic types $\mathrm{NP}$ and $\mathrm{Adv}$. While $\mathrm{Degree}$ is a non-core FE in $\mathrm{Desiring}$ as well as in many other frames, it is a core FE, for instance, in $\mathrm{Control}$, therefore it appears both with and without the prefix $\mathrm{Opt}$ in the full list, if both core and non-core FEs are being considered.

Also note that in GF, the type $\mathrm{Adv}$ (adverbial modifier) covers both adverbs and prepositional phrases (PP), and there is no separate type for PPs.

{\bf The frame module} declares frame valence patterns as functions (henceforth called frame functions) that take one or more FEs and one verb as arguments. These are combined into a clause.

Function names include pattern numbers as suffixes because is is often necessary to separate different valence patterns of the same frame into different functions, for example, in cases when some FEs are mutually exclusive. Moreover, we differentiate functions that return clauses in the passive voice from functions that return active voice clauses because FEs that are subjects in the active voice become objects in the passive voice and vice versa.\footnote{In highly inflected languages, the syntactic function would not change, but this would be reflected by a different word order.} Some examples of frame function signatures regarding the frame $\mathrm{Desiring}$ are listed below. These functions represent common valence patterns found in both BFN and SweFN:

\begin{enumerate}
\item[] $\mathit{fun}$ $\mathrm{Desiring\_P1\_Act :}$\\$\mathrm{Focal\_participant\_NP \to Experiencer\_NP \to}$\\$\mathrm{V2 \to Clause}$
\item[] $\mathit{fun}$ $\mathrm{Desiring\_P1\_Pass :}$\\$\mathrm{Focal\_participant\_NP \to Experiencer\_NP \to}$\\$\mathrm{V2 \to Clause}$
\item[] $\mathit{fun}$ $\mathrm{Desiring\_P2 :}$\\$\mathrm{Event\_VP \to Experiencer\_NP \to V2 \to Clause}$
\end{enumerate}

Each function takes either an intransitive, a transitive or a ditransitive target verb as an argument of type $\mathrm{V}$, $\mathrm{V2}$ or $\mathrm{V3}$ respectively. The required type, in general, depends on the valence pattern. Verbs are defined in the language-specific LU modules.

Frame functions return objects of type $\mathrm{Clause}$ that differs form the type $\mathrm{Cl}$ in RGL. $\mathrm{Clause}$ is a record type consisting of two fields -- $\mathrm{\{np : NP ; vp : VP\}}$ -- whose types are RGL types. It is is a deconstructed $\mathrm{Cl}$ where the subject NP is separated from the rest of the clause. The motivation for this is to allow for nested frames and to allow for adding extra modifiers to the VP before making a $\mathrm{Cl}$ object of the NP and VP parts. An example of nested frames would be the use of the function $\mathrm{Desiring\_P2}$: the FE $\mathrm{Event\_VP}$ would be realized by another frame (e.g. by the frame $\mathrm{Motion}$ in a sentence expressing that someone wants to go somewhere), but only the VP part of the nested frame would be used to fill $\mathrm{Event\_VP}$.

Note that the word order is not considered in the abstract syntax, therefore FEs in the function type signatures are given in the alphabetical order (c.f. Tables~\ref{tab:feBFN} and~\ref{tab:feSweFN}). The target verb is always given as the last argument. The language-specific word order is specified in the concrete syntaxes.

{\bf The LU module} declares a verb sense inventory. Lexemes and their inflectional paradigms for verb senses are specified in the concrete syntaxes. The concrete lexicons also specify valences for the subject and/or object FEs that depend on the verb.

In BFN and SweFN, different approaches are used to identify LUs. A theoretically simple way to unify and align the lexicons via the abstract syntax would be by using English as a meta-language and by including frame names in LU meta-names as sense identifiers. However, LUs between BFN and SweFN are not directly aligned. Although there have been some research on aligning LUs in BFN and SweFN with synsets in Princeton's WordNet \cite{FerrandezEtAl2010,BorinAndForsberg2014}, such indirect alignment currently does not cover many translation equivalents between BFN and SweFN, and among other framenets.

Aligning LUs between framenets is a separate issue which is not addressed in this paper, therefore a framenet-specific, monolingual abstract LU module is currently generated for each language.

Each LU is represented as a function that takes no arguments and returns a verb of type $\mathrm{V}$, $\mathrm{V2}$ or $\mathrm{V3}$. To distinguish between different verb senses, we add the name of the frame to each LU function name, e.g.:

\begin{enumerate}[noitemsep]
\item[] $\mathit{fun}$ $\mathrm{go\_V\_Motion : V}$
\item[] $\mathit{fun}$ $\mathrm{go\_V\_Becoming : V}$
\end{enumerate}

Note that function names do not include framenet-specific sense identifiers, however distinguishing part of speech (POS) tags which correspond to RGL types are still included despite the different value types:\footnote{The nominal LU is given just for an illustration. Compare this to the example LUs given in Sections~\ref{ssec:bfn} and~\ref{ssec:Swefn}.}

\begin{enumerate}[noitemsep]
\item[] $\mathit{fun}$ $\mathrm{\mathring{a}tr\mathring{a}\_V2\_Desiring : V2}$
\item[] $\mathit{fun}$ $\mathrm{\mathring{a}tr\mathring{a}\_N\_Desiring : N}$
\end{enumerate}

We assume that LUs evoking the same frame use a shared set of syntactic valence patterns, but not necessarily every LU uses every valence pattern. An alternative approach would be making valence patterns LU-specific by merging the frame module into the LU module. This would also allow for eliminating the LU argument in the frame building functions, limiting the possible LU-frame mappings.

In the long term, it would be preferable to switch to the more precise LU-valence patterns but, again, LUs between BFN and SweFN and among other framenets are not directly aligned, and the lexicons vary in terms of coverage depending on the underlying corpora. Altogether, this would make it difficult to provide a wide coverage shared abstract syntax.

\subsection{Automatic extraction scenarios}
\label{ssec:fn-gf}

There are several decisions that have to be made in the automatic extraction of a FrameNet-based grammar, more specifically, the extraction of a semantic resource grammar library built on top of the syntactic RGL:
\begin{enumerate}
\item Provide one function per frame and per grammatical voice that takes all core and possibly non-core FEs as arguments so that any argument is optional, i.e. can be an empty phrase, or provide several functions per frame -- one for each typical valence pattern extracted from a corpus.
\item Include non-core FEs as arguments in the frame functions, or do not include.
\item Keep only the intersection of valence patterns (frame functions) extracted from different framenets, or keep the union.
\end{enumerate}

In this paper:
\begin{enumerate}
\item We split the set of core FEs into two or more valence patterns (alternative frame functions) according to the corpus evidence. It is often practically impossible or uncommon that all core FEs could be used in the same sentence. For instance, $\mathrm{Area}$ is mutually exclusive with five other core FEs in the frame $\mathrm{Motion}$, and these five other adverbial modifier FEs normally are not used altogether as well.
\item We do not consider the inclusion of non-core FEs because it is the core FEs that make the frame unique and different from other frames \cite{RuppenhoferEtAl2010}. For adding non-core FEs (adjuncts), the application grammar developer will have to fall back to the API of the syntactic RGL (see the design motivation for $\mathrm{Clause}$ in Section~\ref{ssec:gf-abstract}).
\item We consider only the shared valence patterns between BFN and SweFN as we are primarily interested in multilingual applications and, thus, in functions whose implementation can be generated for all languages.
\end{enumerate}

Although currently we focus only on the extraction of the abstract syntax of a FrameNet-based grammar, the uniform intermediate data that is extracted for generating the abstract syntax contains sufficient information for generating the concrete syntaxes afterwards.

\section{Extracting the abstract syntax from FrameNet-annotated corpora}
\label{sec:experiments}

\subsection{Sentence patterns}
\label{ssec:sentences}

The first step in the extraction of common valence patterns for a multilingual grammar is to convert the FrameNet-annotated sentences into more general and uniform sentence patterns. By sentence patterns we mean sequences of FEs that are subcategorized by the interlingual RGL types (see the FE module in Section~\ref{ssec:gf-abstract}).

There is no unified annotation model used across framenets. BFN and SweFN use not only different XML schemas and POS tagsets; they also use different approaches for annotating the syntactic structure of a sentence.

In BFN, a phrase-structure approach has been taken for specifying a phrase type (PT) for each FE as a whole. Shallow grammatical functions (GF) of FEs are specified as well: an external argument (a phrase outside the VP of the target verb) which is typically the subject; the first object, either direct or indirect (in the case of the active voice); a general dependent.

In SweFN, a dependency approach has been taken: all grammatical annotations, i.e. morpho-syntactic descriptions and dependency relations, are specified at the word level, and no PTs are specified for whole FEs.

A simplified excerpt from the BFN corpus for the LU \emph{want.v} evoking the frame $\mathrm{Desiring}$ is:

\begingroup
\fontsize{7.5pt}{9.5pt}\selectfont
\begin{verbatim}
<sentence ID="732945">
 <text>Traders in the city want a change.</text>
 <annotationSet>
  <layer rank="1" name="BNC">
   <label start="0" end="6" name="NP0"/>
   <label start="20" end="23" name="VVB"/>
   <label start="25" end="25" name="AT0"/>
  </layer>
 </annotationSet>
 <annotationSet status="MANUAL">
  <layer rank="1" name="FE">
   <label start="0" end="18" name="Experiencer"/>
   <label start="25" end="32" name="Event"/>
  </layer>
  <layer rank="1" name="GF">
   <label start="0" end="18" name="Ext"/>
   <label start="25" end="32" name="Obj"/>
  </layer>
  <layer rank="1" name="PT">
   <label start="0" end="18" name="NP"/>
   <label start="25" end="32" name="NP"/>
  </layer>
  <layer rank="1" name="Target">
   <label start="20" end="23" name="Target"/>
  </layer>
 </annotationSet>
</sentence>
\end{verbatim}
\endgroup

A simplified excerpt from the SweFN corpus for the verb \emph{vilja.vb} that evokes the frame $Desiring$ is:\footnote{A description of POS, MSD and dependency tags used in SweFN is available here: \url{http://stp.lingfil.uu.se/~nivre/swedish_treebank/}}

\begingroup
\fontsize{7.5pt}{9.5pt}\selectfont
\begin{verbatim}
<sentence id="ebca5af9-e0494c4e">
 <w pos="JJ" ref="1" dephead="2" deprel="DT">
  Nästa
 </w>
 <w pos="NN" ref="2" dephead="3" deprel="TA">
  gång
 </w>
 <w pos="VB" ref="3" deprel="ROOT">
  skulle
 </w>
 <element name="Experiencer">
  <w pos="PN" ref="4" dephead="3" deprel="SS">
   jag
  </w>
 </element>
 <element name="LU">
  <w msd="VB.AKT" ref="5" dephead="3" deprel="VG">
   vilja
  </w>
 </element>
 <element name="Event">
  <w msd="VB.INF" ref="6" dephead="5" deprel="VG">
   ha
  </w>
  <w pos="RG" ref="7" dephead="8" deprel="DT">
   sju
  </w>
  <w pos="NN" ref="8" dephead="6" deprel="OO">
   sångare
  </w>
 </element>
</sentence>
\end{verbatim}
\endgroup

For each sentence in both BFN and SweFN, a semi-heuristic detection of the grammatical voice and, consequentially, the subject and/or object FE (if any) is performed. Then for each FE in a sentence, its language-specific grammatical type is generalized into an RGL type.

The following generalization rules are applied in BFN:
\begin{enumerate}[noitemsep]
\item {\tt PP {\scriptsize AND} Obj $\to$ NP}
\item {\tt PP {\scriptsize OR} AVP {\scriptsize OR} AJP $\to$ Adv}
\item {\tt NP {\scriptsize AND} {\scriptsize NOT}(Subj {\scriptsize OR} Obj)) $\to$ Adv}
\item {\tt VPto $\to$ VP}
\end{enumerate}

In SweFN, the generalization is done in a similar way, based on the grammatical annotations of the first constituent of an FE. If the first constituent is and adjective or participle, the head word's dependency relation is considered; if it is a coordinating conjunction, next constituent is analysed instead.

Note that currently we do not consider a number of other PTs found in BFN: other kinds of VP (finite, bare stem, participial, gerundive) and PP (if the preposition governs a wh-interrogative clause or is followed by a gerund object), and all kinds of (sub-)clause PTs  \cite{RuppenhoferEtAl2010}. Sentences containing such FEs are ignored ({\texttildelow}14\%). Similarly in SweFN, if a sentence contains an FE that is realized as a subclause, the sentence is ignored ({\texttildelow}4\%).

Regarding the $\mathrm{VPto}$ type in BFN, currently it is always generalized to $\mathrm{VP}$ assuming that it is the object even though it might be a modifier, e.g. \emph{``want to solve it"} vs. \emph{``work to solve it"}. While it is problematic to distinguish these cases in BFN that follows the phrase-structure approach, this distinction is specified in SweFN that follows the dependency grammar approach. Here we could benefit from the multilingual perspective finding complementary information available in one framenet when processing another.

A characteristic of BFN is that FEs which are missing in the sentence are still annotated if the grammar allows or requires the omission, or the identity/type of an FE is understood from the context \cite{RuppenhoferEtAl2010}. Such FEs would be potentially interesting to consider, however, as they have no grammatical annotations, we ignore them keeping the rest FEs in the sentence pattern.

Some of the extracted sentence patterns from BFN for the frame $\mathrm{Desiring}$ are the following:

\begingroup
\fontsize{8pt}{10pt}\selectfont
\begin{verbatim}
Desiring Act  Event_NP.Obj Experiencer_NP.Subj
Desiring Act  Experiencer_NP.Subj Event_NP.Obj
Desiring Act  Experiencer_NP.Subj Event_NP.Obj
Desiring Act  Experiencer_NP.Subj Event_Adv[for]
Desiring Act  Event_VP
Desiring Act  Experiencer_NP.Subj Event_VP
Desiring Pass Event_NP.Subj Experiencer_NP.Obj
\end{verbatim}
\endgroup

For each accepted corpus example, one line is produced in the output specifying the frame, the grammatical voice and the list of expressed FEs according to their word order in the sentence. In addition to the RGL types, we also specify the syntactic function of each FE if is realized as an NP, and we keep track of prepositions that are used to realize PPs. A reference to the target LU and the sentence identifier is not shown but is included as well.

The same format is used for representing SweFN examples, and the same processor is then run in all the remaining steps, including the generation of the abstract syntax.

\subsection{Experiment series}
\label{ssec:experiments}

The uniform sentence patterns are summarized and grouped into valence patterns ignoring the word order and prepositions. As an example, a partial summary of active voice patterns for the frame $\mathrm{Desiring}$ in BFN is:

\begingroup
\fontsize{8pt}{10pt}\selectfont
\begin{verbatim}
Event_VP  Experiencer_NP.Subj               : 53
  Experiencer_NP.Subj  Event_VP               51
  Event_VP  Experiencer_NP.Subj               2
Event_NP.Obj  Experiencer_NP.Subj           : 38
  Experiencer_NP.Subj  Event_NP.Obj           26
  Event_NP.Obj  Experiencer_NP.Subj           12
Event_Adv  Experiencer_NP.Subj              : 23
  Experiencer_NP.Subj  Event_Adv[for]         20
  Experiencer_NP.Subj  Event_Adv[after]       3
Event_VP                                    : 1
  Event_VP                                    1
\end{verbatim}
\endgroup

The above summary shows the typical semantic and syntactic valence patterns for the frame. For the abstract syntax, we consider only the generalized patterns. The specific sentence patterns, including the word order for linearizing $\mathrm{Adv}$ FEs, will later guide the generation of concrete syntaxes.

To estimate pros and cons of the choices made in Sections~\ref{ssec:fn-gf} and~\ref{ssec:sentences}, we run a series of experiments:

\begin{enumerate}[itemsep=0pt,parsep=0pt]
\item[0.0] Extract sentence patterns using framenet-specific syntactic types and skipping null FEs (baseline 0).
\item[1.0] In addition to 0.0, skip examples containing unconsidered grammatical types (baseline 1).
\begin{enumerate}[itemsep=0pt,parsep=0pt,topsep=0pt,partopsep=0pt]
\item[1.A] Skip repeated FEs, then 1.0.
\item[1.B] Skip non-core FEs, then 1.A.
\end{enumerate}
\item[2.0] In addition to 1.0, generalize grammatical types according to GF RGL (baseline 2).
\begin{enumerate}[itemsep=0pt,parsep=0pt,topsep=0pt,partopsep=0pt]
\item[2.A] Skip repeated FEs, then 2.0.
\item[2.B] Skip non-core FEs, then 2.A.
\end{enumerate}
\item[3.0] In addition to 2.0, skip once-used valence patterns.
\begin{enumerate}[itemsep=0pt,parsep=0pt,topsep=0pt,partopsep=0pt]
\item[3.A] Skip repeated FEs, then 3.0.
\item[3.B] Skip non-core FEs, 3.A.
\end{enumerate}
\end{enumerate}

What concerns repeated FEs, there are {\texttildelow}4,000 examples in BFN and {\textgreater}200 in SweFN where more than one chunk is annotated by the same type of FE, mostly due to coordination, wh-words making discontinuous PPs in questions, and anchors of relative clauses (if a frame is evoked in a relative clause). In Settings~x.A, only one FE of the same type is kept. If repeated FEs are of different RGL types, the whole example is skipped.

Regarding non-core FEs, there are {\textgreater}24,500 instances in BFN examples and {\textgreater}2,400 in SweFN that are skipped in Settings~x.B. This reduces the average number of FEs per pattern from 3 to 2.

The results are summarized in Tables~\ref{bfn_statistics} and~\ref{swefn_statistics}. We are primarily interested in Settings~2.B which are rather optimal for both BFN and SweFN: the number of covered frames slightly decreases but it makes the resulting patterns more prototypical and significantly reduces the number of functions to be generated in the abstract syntax, the API. For a large corpus like BFN, skipping once-used valence patterns would help reducing noise but, for a relatively small corpus like SweFN, it would not be reasonable.

\begin{table}[t]
\tabcolsep 4.5pt
\begin{center}
\begin{tabular}{|c|c|rr|rr|rr|}
\hline
\multicolumn{1}{|c}{\multirow{2}{*}{\rotatebox[origin=c]{90}{Settings~~}}} & \multicolumn{1}{|c}{\multirow{2}{*}{\rotatebox[origin=c]{90}{Frames~~~}}} & \multicolumn{2}{|c}{\begin{tabular}[c]{@{}c@{}}Valence\\[-2pt]patterns\end{tabular}} & \multicolumn{2}{|c}{\begin{tabular}[c]{@{}c@{}}Sentence\\[-2pt]patterns\end{tabular}} & \multicolumn{2}{|c|}{\begin{tabular}[c]{@{}c@{}}Corpus\\[-2pt]examples\end{tabular}}\\
\cline{3-8}
\multicolumn{1}{|c}{} & \multicolumn{1}{|c}{} & \multicolumn{1}{|c}{total} & \multicolumn{1}{|c}{\begin{tabular}[c]{@{}c@{}}per\\[-2pt]frame\end{tabular}} & \multicolumn{1}{|c}{total} & \multicolumn{1}{|c}{\begin{tabular}[c]{@{}c@{}}{\small per}\\[-4pt]{\small val.}\\[-4pt]{\small patt.}\end{tabular}} & \multicolumn{1}{|c}{total} & \multicolumn{1}{|c|}{\begin{tabular}[c]{@{}c@{}}{\small per}\\[-4pt]{\small sent.}\\[-4pt]{\small patt.}\end{tabular}}\\
\hline
{\tt 0.0} & 556 & 20623 & 37 & 26427 & 1.3 & 68577 & 2.6\\
\hline
{\tt 1.0} & 552 & 16932 & 31 & 22424 & 1.3 & 59073 & 2.6\\
{\tt 1.A} & 550 & 14830 & 27 & 20350 & 1.4 & 57902 & 2.8\\
{\tt 1.B} & 550 & 5626 & 10 & 8378 & 1.5 & 58085 & 6.9\\
\hline
{\tt 2.0} & 552 & 14811 & 27 & 22191 & 1.5 & 59073 & 2.7\\
{\tt 2.A} & 550 & 12799 & 23 & 20286 & 1.6 & 58423 & 2.9\\
{\bfseries\ttfamily 2.B} & {\bf 550} & {\bf 5079} & {\bf 9} & {\bf 8339} & {\bf 1.6} & {\bf 58431} & {\bf 7.0}\\
\hline
{\tt 3.0} & 500 & 5835 & 12 & 13215 & 2.3 & 50097 & 3.8\\
{\tt 3.A} & 498 & 5498 & 11 & 12985 & 2.4 & 51122 & 3.9\\
{\tt 3.B} & 503 & 3277 & 7 & 6537 & 2.0 & 56629 & 8.7\\
\hline
\end{tabular}
\caption{Frame-specific pattern extraction from BFN. Sentence patterns preserve the word order and prepositions. Valence patterns disregard both.}
\label{bfn_statistics}
\end{center}
\end{table}

\begin{table}[t]
\tabcolsep 4.5pt
\begin{center}
\begin{tabular}{|c|c|rr|rr|rr|}
\hline
\multicolumn{1}{|c}{\multirow{2}{*}{\rotatebox[origin=c]{90}{Settings~~}}} & \multicolumn{1}{|c}{\multirow{2}{*}{\rotatebox[origin=c]{90}{Frames~~~}}} & \multicolumn{2}{|c}{\begin{tabular}[c]{@{}c@{}}Valence\\[-2pt]patterns\end{tabular}} & \multicolumn{2}{|c}{\begin{tabular}[c]{@{}c@{}}Sentence\\[-2pt]patterns\end{tabular}} & \multicolumn{2}{|c|}{\begin{tabular}[c]{@{}c@{}}Corpus\\[-2pt]examples\end{tabular}}\\
\cline{3-8}
\multicolumn{1}{|c}{} & \multicolumn{1}{|c}{} & \multicolumn{1}{|c}{total} & \multicolumn{1}{|c}{\begin{tabular}[c]{@{}c@{}}per\\[-2pt]frame\end{tabular}} & \multicolumn{1}{|c}{total} & \multicolumn{1}{|c}{\begin{tabular}[c]{@{}c@{}}{\small per}\\[-4pt]{\small val.}\\[-4pt]{\small patt.}\end{tabular}} & \multicolumn{1}{|c}{total} & \multicolumn{1}{|c|}{\begin{tabular}[c]{@{}c@{}}{\small per}\\[-4pt]{\small sent.}\\[-4pt]{\small patt.}\end{tabular}}\\
\hline
{\tt 0.0} & 638 & ~~3404 & 5.3 & ~~3435 & 1.0 & ~~3697 & 1.1\\
\hline
{\tt 1.0} & 636 & 3269 & 5.1 & 3300 & 1.0 & 3546 & 1.1\\
{\tt 1.A} & 627 & 3122 & 5.0 & 3153 & 1.0 & 3359 & 1.1\\
{\tt 1.B} & 629 & 2759 & 4.4 & 2813 & 1.0 & 3398 & 1.2\\
\hline
{\tt 2.0} & 636 & 2829 & 4.4 & 2967 & 1.0 & 3543 & 1.2\\
{\tt 2.A} & 632 & 2729 & 4.3 & 2877 & 1.1 & 3438 & 1.2\\
{\bfseries\ttfamily 2.B} & {\bf 632} & {\bf 1866} & {\bf 3.0} & {\bf 2029} & {\bf 1.1} & {\bf 3460} & {\bf 1.7}\\
\hline
{\tt 3.0} & 308 & 472 & 1.5 & 610 & 1.3 & 1186 & 1.9\\
{\tt 3.A} & 305 & 465 & 1.5 & 613 & 1.3 & 1174 & 1.9\\
{\tt 3.B} & 462 & 714 & 1.5 & 877 & 1.2 & 2308 & 2.6\\
\hline
\end{tabular}
\caption{Frame-specific pattern extraction from SweFN.}
\label{swefn_statistics}
\end{center}
\end{table}

\begin{table*}[ht]
\tabcolsep 4pt
\begin{center}
\begin{tabular}{|c|c|c|rr|rr|c|rr|}
\hline
\multicolumn{1}{|c}{\multirow{2}{*}{\begin{tabular}[c]{@{}c@{}}Settings\\{\tt BFN:SweFN}\end{tabular}}} & \multicolumn{9}{|c|}{Frames}\\
\cline{2-10}
\multicolumn{1}{|c}{} & \multicolumn{1}{|c}{{\tt BFN}} & \multicolumn{1}{|c}{{\tt SweFN}} & \multicolumn{2}{|c}{{\tt BFN$\setminus$SweFN}} & \multicolumn{2}{|c}{{\tt SweFN$\setminus$BFN}} & \multicolumn{1}{|c}{{\tt BFN$\cup$SweFN}} & \multicolumn{2}{|c|}{{\tt BFN$\cap$SweFN}}\\
\hline
{\bfseries\ttfamily 2.B:2.B} & {\bf 550} & {\bf 632} & {\bf 57} & ({\bf 10}\%) & {\bf 139} & ({\bf 22}\%) & {\bf 689} & {\bf 493} & ({\bf 72}\%)\\
{\tt 3.B:2.B} & 503 & 632 & 46 & (9\%) & 175 & (28\%) & 678 & 457 & (67\%)\\
\hline
\end{tabular}
\caption{Comparison of frames in BFN and SweFN. Symbols $\setminus$, $\cup$ and $\cap$ denote the set operations \emph{difference}, \emph{union} and \emph{intersection}. E.g. comparing BFN~2.B and SweFN~2.B, for 139 frames there are verb-evoked examples only in SweFN.}
\label{tab:frames}
\end{center}
\end{table*}

\begin{table*}[ht]
\tabcolsep 4pt
\begin{center}
\begin{tabular}{|c|c|c|rr|rr|c|rr|c|c|}
\hline
\multicolumn{1}{|c}{\multirow{2}{*}{\begin{tabular}[c]{@{}c@{}}Settings\\{\tt BFN:SweFN}\end{tabular}}} & \multicolumn{9}{|c|}{Semantic valence patterns} & \multicolumn{2}{|c|}{Final}\\
\cline{2-12}
\multicolumn{1}{|c}{} & \multicolumn{1}{|c}{{\tt BFN}} & \multicolumn{1}{|c}{{\tt SweFN}} & \multicolumn{2}{|c}{{\tt BFN$\setminus$SweFN}} & \multicolumn{2}{|c}{{\tt SweFN$\setminus$BFN}} & \multicolumn{1}{|c}{{\tt BFN$\cup$SweFN}} & \multicolumn{2}{|c|}{{\tt BFN$\cap$SweFN}} & Patterns & Frames\\
\hline
{\bfseries\ttfamily 2.B:2.B} & {\bf 2374} & {\bf 1210} & {\bf 1418} & ({\bf 60}\%) & {\bf 254} & ({\bf 21}\%) & {\bf 2628} & {\bf 956} & ({\bf 36}\%) & {\bf 615} & {\bf 466}\\
{\tt 3.B:2.B} & 1768 & 1123 & 932 & (53\%) & 287 & (26\%) & 2055 & 836 & (41\%) & 551 & 433\\
\hline
\multicolumn{12}{c}{~}\\
\hline
\multicolumn{1}{|c}{\multirow{2}{*}{\begin{tabular}[c]{@{}c@{}}Settings\\{\tt BFN:SweFN}\end{tabular}}} & \multicolumn{9}{|c|}{Semantic-syntactic valence patterns} & \multicolumn{2}{|c|}{Final}\\
\cline{2-12}
\multicolumn{1}{|c}{} & \multicolumn{1}{|c}{{\tt BFN}} & \multicolumn{1}{|c}{{\tt SweFN}} & \multicolumn{2}{|c}{{\tt BFN$\setminus$SweFN}} & \multicolumn{2}{|c}{{\tt SweFN$\setminus$BFN}} & \multicolumn{1}{|c}{{\tt BFN$\cup$SweFN}} & \multicolumn{2}{|c|}{{\tt BFN$\cap$SweFN}} & Patterns & Frames\\
\hline
{\bfseries\ttfamily 2.B:2.B} & {\bf 3212} & {\bf 1446} & {\bf 2279} & ({\bf 71}\%) & {\bf 513} & ({\bf 35}\%) & {\bf 3725} & {\bf 933} & ({\bf 25}\%) & {\bf 655} & {\bf 440}\\
{\tt 3.B:2.B} & 2184 & 1344 & 1374 & (63\%) & 534 & (40\%) & 2718 & 810 & (30\%) & 575 & 408\\
\hline
\end{tabular}
\caption{Comparison of valence patterns in BFN and SweFN: by the exact match. Compare with Table~\ref{tab:patterns_fuzzy}.}
\label{tab:patterns_exact}
\end{center}
\end{table*}

\begin{table*}[t]
\tabcolsep 4pt
\begin{center}
\begin{tabular}{|c|r|r|rr|rr|c|rr|c|c|}
\hline
\multicolumn{1}{|c}{\multirow{2}{*}{\begin{tabular}[c]{@{}c@{}}Settings\\{\tt BFN:SweFN}\end{tabular}}} & \multicolumn{9}{|c|}{Semantic valence patterns} & \multicolumn{2}{|c|}{Final}\\
\cline{2-12}
\multicolumn{1}{|c}{} & \multicolumn{1}{|c}{{\tt BFN}} & \multicolumn{1}{|c}{{\tt SweFN}} & \multicolumn{2}{|c}{{\tt BFN$\setminus$SweFN}} & \multicolumn{2}{|c}{{\tt SweFN$\setminus$BFN}} & \multicolumn{1}{|c}{{\tt BFN$\cup$SweFN}} & \multicolumn{2}{|c|}{{\tt BFN$\cap$SweFN}} & Patterns & Frames\\
\hline
{\bfseries\ttfamily 2.B:2.B} & {\bf 2374} & {\bf 1210} & {\bf 842} & ({\bf 35}\%) & {\bf 142} & ({\bf 12}\%) & {\bf 2628} & {\bf 1644} & ({\bf 63}\%) & {\bf 678} & {\bf 488}\\
{\tt 3.B:2.B} & 1768 & 1123 & 515 & (29\%) & 165 & (15\%) & 2055 & 1375 & (67\%) & 610 & 451\\
\hline
\multicolumn{12}{c}{~}\\
\hline
\multicolumn{1}{|c}{\multirow{2}{*}{\begin{tabular}[c]{@{}c@{}}Settings\\{\tt BFN:SweFN}\end{tabular}}} & \multicolumn{9}{|c|}{Semantic-syntactic valence patterns} & \multicolumn{2}{|c|}{Final}\\
\cline{2-12}
\multicolumn{1}{|c}{} & \multicolumn{1}{|c}{{\tt BFN}} & \multicolumn{1}{|c}{{\tt SweFN}} & \multicolumn{2}{|c}{{\tt BFN$\setminus$SweFN}} & \multicolumn{2}{|c}{{\tt SweFN$\setminus$BFN}} & \multicolumn{1}{|c}{{\tt BFN$\cup$SweFN}} & \multicolumn{2}{|c|}{{\tt BFN$\cap$SweFN}} & Patterns & Frames\\
\hline
{\bfseries\ttfamily 2.B:2.B} & {\bf 3212} & {\bf 1446} & {\bf 1693} & ({\bf 53}\%) & {\bf 402} & ({\bf 28}\%) & {\bf 3725} & {\bf 1630} & ({\bf 44}\%) & {\bf 769} & {\bf 471}\\
{\tt 3.B:2.B} & 2184 & 1344 & 955 & (44\%) & 415 & (31\%) & 2718 & 1348 & (50\%) & 674 & 435\\
\hline
\end{tabular}
\caption{Comparison of valence patterns in BFN and SweFN: by the fuzzy match. Compare with Table~\ref{tab:patterns_exact}.}
\label{tab:patterns_fuzzy}
\end{center}
\end{table*}

\begin{table*}[t!]
\tabcolsep 4pt
\begin{center}
\begin{tabular}{|c|c|c|cc|cc|c|cc|cc|}
\hline
\multirow{2}{*}{Types} & \multirow{2}{*}{Match} & \multicolumn{5}{|c}{BFN} & \multicolumn{5}{|c|}{SweFN}\\
\cline{3-12}
&& \multicolumn{3}{|c}{By patterns} & \multicolumn{2}{|c}{By frames} & \multicolumn{3}{|c}{By patterns} & \multicolumn{2}{|c|}{By frames}\\
\hline
\multirow{2}{*}{Semantic}
& Exact       & 48351       & 87.7\%       & (81.8\%)       & 55111       & (93.3\%)       & 2533       & 92.4\%       & (71.5\%)       & 2742       & (77.4\%)\\
& {\bf Fuzzy} & {\bf 48831} & {\bf 88.0}\% & ({\bf 82.7}\%) & {\bf 55518} & ({\bf 94.0}\%) & {\bf 2602} & {\bf 92.5}\% & ({\bf 73.4}\%) & {\bf 2812} & ({\bf 79.4}\%)\\
\hline
\multirow{2}{*}{Syntactic}
& Exact        & 41464      & 76.4\%       & (70.2\%)       & 54259       & (91.9\%)       & 1993       & 75.2\%       & (56.3\%)       & 2652       & (74.9\%)\\
& {\bf Fuzzy} & {\bf 42897} & {\bf 77.8}\% & ({\bf 72.6}\%) & {\bf 55166} & ({\bf 93.4}\%) & {\bf 2069} & {\bf 74.9}\% & ({\bf 58.4}\%) & {\bf 2763} & ({\bf 78.0}\%)\\
\hline
\end{tabular}
\caption{The proportion of examples in BFN and SweFN whose sentence patterns belong to the final set of shared frames and are covered by the final set of shared semantic / semantic-syntactic valence patterns (Tables~\ref{tab:patterns_exact} and~\ref{tab:patterns_fuzzy}, Settings~2.B:2.B). The ratio relative to all examples is given in parenthesis. Coverage considering only frames is given additionally.}
\label{tab:patterns_coverage}
\end{center}
\end{table*}

\subsection{Shared valence patterns}
\label{ssec:valences}

To find a set of valence patterns that are shared between BFN and SweFN, we compared the outcome of both framenets in Settings~2.B. Additionally, we included Settings~3.B for BFN. It turns out that the pair of BFN~3.B and SweFN~2.B produces a proper subset of frames / valence patterns if compared to BFN~2.B and SweFN~2.B.

The comparison is first done for the sets of verb frames (Table~\ref{tab:frames}) and then for the sets of valence patterns that belong to the shared set of frames (Tables~\ref{tab:patterns_exact} and~\ref{tab:patterns_fuzzy}).

Valence patterns are compared at two levels: considering only the semantic types of FEs and considering both the semantic and the syntactic types. The difference indicates the variation due to alternative syntactic realizations in terms of the RGL types. For example, the following semantic valence pattern of the frame $\mathrm{Desiring}$ is realized by two shared semantic-syntactic patterns:

\begingroup
\fontsize{8.5pt}{10.5pt}\selectfont
\begin{verbatim}
Experiencer Focal_participant
  Experiencer_NP Focal_participant_Adv
  Experiencer_NP Focal_participant_NP
\end{verbatim}
\endgroup

Valence patterns are compared in two ways: by exact match (Table~\ref{tab:patterns_exact}) and by subsumption, a fuzzy match (Table~\ref{tab:patterns_fuzzy}). Pattern $\mathrm{A}$ subsumes pattern $\mathrm{B}$ if: $\mathrm{A.frame = B.frame}$, $\mathrm{A.voice = B.voice}$ (if comparing syntactic types), and $\mathrm{B.FEs \subseteq A.FEs}$. If a pattern of $\mathrm{Framenet_1}$ is subsumed by a pattern of $\mathrm{Framenet_2}$, it is added to the intersection (and vice versa). In the final intersection, patterns which are subsumed by other patterns are removed. This approach is backed up by the design of the semantic grammar which accepts an empty phrase as an argument to a frame function if the corresponding FE is not expressed in the sentence.

The fuzzy match results in 10\% more patterns that cover 5\% more frames if comparing only the semantic types, and 17\% more patterns that cover 7\% more frames if comparing the semantic-syntactic types.

For the frame $\mathrm{Desiring}$, there are three shared valence patterns included in the final abstract syntax:

\begingroup
\fontsize{8.5pt}{10.5pt}\selectfont
\begin{verbatim}
Event_VP Experiencer_NP
Experiencer_NP Focal_participant_Adv
Experiencer_NP Focal_participant_NP
\end{verbatim}
\endgroup

This covers all patterns found in SweFN, however, several patterns found only in BFN are not included, like:

\begingroup
\fontsize{8.5pt}{10.5pt}\selectfont
\begin{verbatim}
Event_Adv Experiencer_NP Focal_participant_NP
Event_NP Experiencer_NP
Event_VP Experiencer_NP Focal_participant_NP
\end{verbatim}
\endgroup

The shared FE module declares 423 semantic-syntactic types: 291 of type $\mathrm{NP}$, 126 of type $\mathrm{Adv}$ and 6 of type $\mathrm{VP}$. If considering only semantic types, there are 349.


\subsection{Evaluation}
\label{ssec:evaluation}

We evaluate the resulting abstract syntax by counting the number of examples in BFN and SweFN whose sentence patterns belong to the final set of shared frames and are covered by the final set of shared semantic-syntactic valence patterns. The final sets are extracted applying Settings 2.B for both framenets using both the exact and the fuzzy comparison alternatively. Corpus examples are represented by sentence patterns according to Settings~2.0 disregarding repeated and non-core FEs, word order and prepositions.

Additionally, we measure coverage by the shared semantic patterns (disregarding syntactic types) to estimate the upper limit that could be reached by expanding the syntactic variation in the final set of semantic-syntactic patterns, introducing no additional semantic pattern. To estimate the upper limit that could be reached by expanding the semantic variation in the set of semantic patterns, we measure coverage by the shared set of frames (disregarding valences).

The results are summarized in Table~\ref{tab:patterns_coverage}. With a total number of just 769 valence patterns (frame function signatures), the abstract syntax covers 77.8\% of BFN examples that belong to the shared set of 471 frames (72.6\% of all BFN examples). For SweFN, the ratio is 74.9\% and 58.4\% respectively. The huge drop of coverage relative to all examples in SweFN is due to many frames that have verb-evoked examples only in SweFN (Table~\ref{tab:frames}).

Coverage by the semantic patterns is {\texttildelow}10\% higher for BFN and {\texttildelow}15--18\% higher for SweFN. The upper limit that could be reached by the shared set of frames (considering all examples) is 93.4\% for BFN and 78\% for SweFN. This shows that the extracted abstract syntax covers common frames and valence patterns. Moreover, the most frequent frames and patterns are already included by the exact match; the fuzzy match gives only a slight improvement.

\section{Conclusion}
\label{sec:conclusions}

We have presented the extraction of the abstract syntax of currently bilingual but potentially multilingual FrameNet-based grammar. The acquired abstract syntax is compact but comprehensive, and the initial results are very promising. Despite the relatively small SweFN corpus, the shared semantic and syntactic valence patterns cover most examples in both corpora. The resulting grammar approximates the actual grammars and, in general, examples are covered by paraphrasing. By including more framenets into comparison, the intersection among framenets would decrease but the resulting patterns should become more accurate and prototypical. We have also observed that the grammar would not expand much if common non-core FEs were added.

Our method for comparing framenets at the grammar level provides additional means for checking quality, consistency and coverage of a FrameNet-annotated corpus; we logged lots of ill-annotated examples in both BFN and SweFN.

While focusing on the automatic extraction of the grammar, the concise outcome allows for manual adjustments afterwards, such as adjusting the few $\mathrm{VP}$ and $\mathrm{Adv}$ cases.

The next task is to generate concrete syntaxes for English and Swedish. We also intend to include more languages and more syntactic types, such as subclauses, as well as consider verbal nouns as target LUs.

\section{Acknowledgements}
This research has been supported by the Swedish Research Council under grant No. 2012-5746 (Reliable Multilingual Digital Communication: Methods and Applications), by the Centre for Language Technology in Gothenburg, and by the Latvian National Research Programme in Information Technology (project No.~5).

\bibliographystyle{lrec2014}
\bibliography{lrecBib}

\end{document}